\documentclass{article}
\usepackage{amsmath}
\usepackage{arxiv}

\usepackage[utf8]{inputenc} 
\usepackage[T1]{fontenc}    
\usepackage{hyperref}       
\usepackage{url}            
\usepackage{booktabs}       
\usepackage{amsfonts}       
\usepackage{nicefrac}       
\usepackage{microtype}      
\usepackage{cleveref}       
\usepackage{lipsum}         
\usepackage{graphicx}
\usepackage{natbib}
\usepackage{doi}
\usepackage{algorithm}
\usepackage{algorithmic}
\usepackage{multirow}
\usepackage{color}
\usepackage{amsbsy}
\usepackage[mathscr]{euscript}
\usepackage{bm}
\usepackage{graphicx,amssymb,mathrsfs,amsfonts,dsfont}
\usepackage{hyperref}
\usepackage{lineno}
\usepackage{subfigure}
\usepackage{stfloats}
\usepackage{algorithm}
\usepackage{algorithmic}
\usepackage{amsbsy}
\usepackage[mathscr]{euscript}
\usepackage{bm}
\newcommand{\x}{\bm{x}}
\newcommand{\y}{\bm{y}}
\newcommand{\n}{\bm{n}}
\newcommand{\w}{\bm{w}}
\newcommand{\z}{\bm{z}}
\newtheorem{thm}{Theorem}[section]

\newtheorem{rmk}{Remark}[section]
\newtheorem{assump}{Assumption}
\title{Equilibrated Zeroth-Order Unrolled Deep Networks for Accelerated MRI}

\author{Zhuo-Xu~Cui \\ SIAT \\
\texttt{zx.cui@siat.ac.cn}  \And Jing~Cheng  \\ SIAT \\ 	
\texttt{jing.cheng@siat.ac.cn} \And Qingyong~Zhu \\ SIAT \\ \texttt{	
qy.zhu@siat.ac.cn}  \And Yuanyuan~Liu 	
\\National Innovation Center for Advanced Medical Devices\\	
\texttt{liuyy@siat.ac.cn} \And Sen~Jia 	
\\ SIAT \\
\texttt{sen.jia@siat.ac.cn} \\ \And Kankan~Zhao
\\ SIAT \\ \texttt{kk.zhao@siat.ac.cn} \And Ziwen~Ke
\\ Shanghai Jiaotong University \\ \texttt{zw.ke@siat.ac.cn} \And Wenqi~Huang
\\ Technical University of Munich \\\texttt{wq.huang1@siat.ac.cn} \And Haifeng~Wang\\SIAT \\	
\texttt{hf.wang1@siat.ac.cn} \And Yanjie~Zhu\\ SIAT \\ 	
\texttt{yj.zhu@siat.ac.cn} \And Dong~Liang 	\\ SIAT \\Corresponding author:\\
\texttt{dong.liang@siat.ac.cn}\\
	\\
}

\renewcommand{\shorttitle}{\textit{arXiv} Template}

\begin{document}
\maketitle

\begin{abstract}
Recently, model-driven deep learning unrolls a certain iterative algorithm of a regularization model into a cascade network by replacing the first-order information (i.e., (sub)gradient or proximal operator) of the regularizer with a network module, which appears more explainable and predictable compared to common data-driven networks.
Conversely, in theory, there is not necessarily such a functional regularizer whose first-order information matches the replaced network module, which means the network output may not be covered by the original regularization model. Moreover, up to now, there is also no theory to guarantee the global convergence and robustness (regularity) of unrolled networks under realistic assumptions.
To bridge this gap, this paper propose to present a safeguarded methodology on network unrolling. Specifically, focusing on accelerated MRI, we unroll a zeroth-order algorithm, of which the network module represents the regularizer itself, so that the network output can be still covered by the regularization model.
Furthermore, inspired by the ideal of deep equilibrium models, before backpropagating, we carry out the unrolled iterative network to converge to a fixed point to ensure the convergence.
In case the measurement data contains noise, we prove that the proposed network is robust against noisy interferences. Finally, numerical experiments show that the proposed network consistently outperforms the state-of-the-art MRI reconstruction methods including traditional regularization methods and other deep learning methods.
\end{abstract}

\keywords{Deep Equilibrium networks, zeroth-order algorithm, accelerated MRI, inverse problem, convergence, regularity.}

\section{Introduction}
MRI is one of the most widely used imaging modality in routine clinical practice, due to its noninvasion, nonionization and excellent visualization of soft-tissue contrast.
But it has been traditionally limited by its slow data acquisition speed. How to shorten the imaging time has become a research hotspot. Particularly,
reconstructing high quality MR image or full-sampled $k$-space data from under-sampled $k$-space data is a direct and effective way to improve the imaging speed \cite{liang1992constrained}.
One of the most successful technical method over the past 20 years was parallel imaging (PI), which mainly capitalizes the multi-coil information to interpolate the under-sampled $k$-space data \cite{sodickson1997simultaneous,Pruessmann1999SENSE}.
In 2006, another high-profile method, termed compressed sensing (CS), was proposed \cite{Candes2006Robust,Donoho2006Compressed}, which made it possible to recover the under-sampled sparse signal completely. Leveraging the sparse nature of MR images, CS has been successfully applied to accelerated MRI (CS-MRI) \cite{Lustig2007Sparse}. Nowadays, most of the state-of-the-art clinical MRI scanners are equipped with PI technology. CS and PI often appear in a combination form for modern accelerated MR reconstruction methods \cite{Liang2009Accelerating,She2014Sparse,7163966,8017620}. For such methods, however, it usually takes a relatively long time to find a high quality solution. Additionally, it can only exploit the universal priors and lacks adaptability for different types of MR data.

Recently, inspired by the demonstrated tremendous success of deep learning (DL), many researches have committed to applied DL to MR reconstruction (termed DL-MRI) and received significant performance gain \cite{7493320,8962949,Huang2021Deep,9481093,9481108,9632354}. DL-MRI adaptively captures the priors in a data-driven manner from training data and performs superfast online reconstruction with the aid of offline training.
Early work on DL-MRI relied mainly on learning mappings between under-sampled $k$-space data (or zero-filling images) and full-sampled $k$-spaces data (or high quality images) \cite{7493320,zhu2018Image}.
Although it achieves excellent results, it has been considered that the reconstruction results may be uncertain due to its separation from MRI model.
Another line of development initiated by \cite{Gregor2010learning} which tarts with iterative algorithms for CS-MRI regularization model and releases the first-order information (i.e., (sub)gradient or proximal operator) of the regularizer and other parameters (including regularization parameter, stepsize, e.t.c.) as learnable to unroll the algorithms to parameterized deep networks \cite{yang2016deep,Zhang_2018_CVPR,8271999,Hammernik2018Learning}. Furthermore, following the line of unrolling, many PI regularization model driven deep networks also has been proposed \cite{akcakaya2019scan,kim2019loraki,9159672}.
Since the architecture of unrolled deep networks (UDN) is driven by CS-MRI or PI regularization models, it appears more explainable and predictable compared to above data-driven networks. A large number of experiments have also confirmed its competitiveness in reconstructed image quality.

In theory, however, the following problems for UDN are still not well studied.
Firstly, for the unrolled network module (i.e., parameterized (sub)gradient or proximal operator), there is not necessarily such a functional regularizer whose first-order information matches it, which means that the output of UDN may not be covered by the original CS-MRI or PI regularization model. In other word, the output of UDN cannot inherit its explainable and predictable nature completely. Although some literatures given the convergence proof of UDN \cite{9152164,Huang2021Deep,9481093}, the corresponding theory requires that the number of iterations tends to be infinite and UDN usually unrolls only a few layers. Therefore, we cannot judge whether the output has converged. Furthermore, \cite{Antun30088} revealed that DL based methods (including UDN) typically yield unstable reconstruction,
which hinders the clinical application of DL methods severely. In short, if one want to promote UDN clinically, at least the following issues need to be addressed:
\begin{enumerate}
	\item Can UDN inherit the explainable and predictable nature of original CS-MRI or PI regularization model completely?
	\item Can the UDN be guaranteed to converge globally?
	\item Is the UDN robust against interferences?
\end{enumerate}

\subsection{Contributions}
Motivated by the questions mentioned above, this paper aims to propose a safeguarded UDN approach for accelerated MR reconstruction. Specifically, the main contributions of the paper are summarized as follows:
\begin{itemize}
	\item Firstly, our new approach starts with a zeroth-order algorithm (projection over convex sets, POCS) of a PI regularization model and releases the zeroth-order information of the regularizer (i.e., regularizer itself) as a parameterized network module. Therefore, there is such a generalized PI regularization model with parameterized functional regularizer that the proposed zeroth-order UDN can be summed up as its solution algorithm. Then, the output of proposed zeroth-order UDN can inherit its explainable and predictable nature well.
	\item  Furthermore, inspired by the ideal of deep equilibrium models (DEQ), before backpropagating, we carry out the zeroth-order UDN to converge to a fixed point to ensure the convergence.
	We also proved that this fixed point enjoys a tight complexity guarantee to approximate the true MR image (full-sampled $k$-space data). In practice, the measurement from different MR system are often mixed with different level noise. We prove that the proposed zeroth-order UDN is robust against noisy interference. What's more, for general UDN networks, previous experience shows that the outcome performance is sensitive to a good initializer, which is difficult to implement in practice. In the proposed method, the network parameters learning is based on the convergent fixed point rather than initial input such that the outcome performance is independent of initial input. In a word, we bridge the theoretical guarantee for UDN based methods in accelerated MRI, in terms of convergence and robustness.
	\item In terms of imaging, the pre-estimated parameter in the regularizer of traditional PI model is absorbed into the network parameters, which are learned from training data rather than a self calibration region, so that the proposed method can enable calibration-free parallel MRI.
	\item Numerical results on two MR datasets with different sampling patterns show that the proposed zeroth-order UDN significantly outperforms traditional PI methods and conventional UDN, as characterized by quantitative metrics and visual evaluation.
\end{itemize}
The remainder of the paper is organized as follows. Section \ref{sect2} provides some notions and preliminaries. Section \ref{sect3} discusses the equilibrated zeroth-order UDN
network for $k$-space PI model. Section \ref{sect4} discusses its corresponding theoretical guarantees. The implementation details are presented in Section \ref{sect5}. Experiments that performed on several data sets are presented in Section \ref{sect6}. The discussions are presented in Section \ref{sect7}. The last section \ref{sect8} gives some concluding remarks. All the proofs are presented in the supplementary material.
\section{Notions}\label{sect2}
In this paper, we let $[N]=\{1,\ldots,N\}$. In the case of no ambiguity, matrices and vectors are all represented by bold lower cases letters, i.e. ${\x}$, $\y$. In addition, ${\x}_{i}$ (${\x}_{i,j}$) refers to the $i$-th column ($(i,j)$th entry) of the matrix ${\x}$.
The superscript ${}^{T}$ for a matrix denotes the transpose and ${}^{*}$ for a operator denotes the adjoint.
A variety of norms on matrices will be discussed. The spectral norm of a matrix ${\x}$ is denoted by $\|{\x}\|$. The Euclidean inner product between two matrices is $\langle{\x},\y\rangle=\text{Trace}({\x}^*\y)$ and the corresponding Euclidean norm, termed Frobenius norm, is denoted as $\|{\x}\|_F$ which is derived by $\|{\x}\|_F:=\langle{\x},{\x}\rangle$. For vectors, $\|\cdot\|$ denotes the $\ell_2$ norm.

We say an operator $T:\mathbb{C}^{N_1}\rightarrow \mathbb{C}^{N_2}$ is $L$-Lipschitz, if it holds
$$T({\x})-T(\y)\leq L\|{\x}-\y\|$$
for any ${\x},\y\in\mathbb{C}^{N_1}$. We say the operator $T$ is nonexpansive if it is 1-Lipschitz. We use $\mathcal{P}_{\mathcal{C}}$ denote the orthogonal projection on $\mathcal{C}$, i.e., $\mathcal{P}_{\mathcal{C}}({\x})=\arg\min_{\y\in \mathcal{C}}\|{\x}-\y\| $. The spatial Fourier transform of arbitrary smooth function ${f}:\mathbb{R}^2\rightarrow\mathbb{R}^1$ is defined by
$$\widehat{f}(\bm{w})=\mathcal{F}[f](\bm{w}):=\int_{\mathbb{R}^d}e^{-i\bm{w}\cdot\bm{r}}f(\bm{r})d\bm{r}$$
with spatial frequency $ \bm{k}\in\mathbb{R}^2$ and $i=\sqrt{-1}$, where $\widehat{\cdot}$ denotes the Fourier transform in short.

\section{Related Works and Methods}\label{sect3}
In this section, at first, we briefly review some related works and then introduce the proposed zeroth-order UDN in detail.
\subsection{Related Works}
\subsubsection{Classical PI methods}
Mathematically, based on multi-coil acquisition, PI model is transformed from a single linear equation to redundant linear equations set.
The task of original SENSE is to solve the redundant equations to reconstruct MR image \cite{pruessmann1999sense1}. With the emergence of CS, SENSE model has also been extend to sparsity regularized form \cite{Knoll2021Parallel}. On the other hand, PI model can also be formulated in the $k$-space as an interpolation procedure, which is achieved by assuming that the values of $k$-space data among each channel are predictable within a neighborhood. Among of them, GRAPPA \cite{Griswold2002Generalized}, SPIRIT \cite{Lustig2010SPIRiT}, e.t.c., are the most prominent examples,
In terms of imaging, compared with SENSE based PI model, since $k$-space PI model is easier to implement without estimating coil sensitivity, it has been widely deployed by commercial MRI scanner vendors.

\subsubsection{Unrolled Deep Networks (UDN)}
The UDN starts with an architecture of an iterative algorithm for CS-MRI or PI regularization problem and releases the first-order information (i.e., (sub)gradient or proximal operator) of the regularizer as learnable convolutional neural network module, i.e., $\text{CNN}_{i}(\cdot)$ at the $i$th layer \cite{yang2016deep,Zhang_2018_CVPR,8271999,Hammernik2018Learning,9152164,9481093,9481108,Huang2021Deep,9159672}. Conversely, there is not necessarily such a functional regularizer whose first-order information matches $\text{CNN}_{i}(\cdot)$. Namely, there is not necessarily such a functional $R$ such that the following equation holds:
$$\text{CNN}_{i}(x)\in \partial R(x) (\text{or}~\text{Prox}_{R}(x)), ~i=1,\ldots,T$$
where $T$ denotes the number of layers of UDN. Therefore, there is no such a CS-MRI or PI regularization model that can cover the output of UDN. In other word, the output of UDN cannot inherit the explainable and predictable nature of original CS-MRI or PI model completely. Although UDN is not theoretically perfect at present, for inverse problem including accelerated MRI, many scholars believe that it presents a potential approach to break the limits of analytic regularization methods \cite{8962949,2021Learning}.

\subsubsection{Deep Equilibrium Models (DEQ)}
In recursive networks (including UDN), it has been found that the deeper the network layers, the stronger the network expressibility. Naturally, one wonders what happens when the number of layers goes to infinity? However, because of the memory limitation, it is intractable for training a network with arbitrarily number layers. Fortunately, recent works on DEQ showed that this limitations can be modeled using a fixed (equilibrium) point equation \cite{2019Deep,2020Multiscale}. Briefly speaking, DEQ firstly executes the recursive networks to converge to a fixed point before backpropagating, which is equivalent to running an infinite depth network. Then, its backpropagation can be analytically computed only through this fixed point, such that the memory does not increase as the depth of the network gets deeper.

DEQ not only realizes the infinite depth on networks, but also sheds light on the convergence of the recursive networks. In theory, based on the monotone operators, \cite{2020Monotone} introduced some efficient solvers for finding fixed points with guaranteed stable convergence. In particular, when the recursive network is designed in a manner of unrolling, \cite{Gilton2021Deep,2021Feasibility} showed that the UDN under the framework of DEQ (dubbed DEQ-UDN) is guaranteed to converge to a fixed point only by a Lipschitz constraint on network parameters.
\subsection{Forward Model}
In accelerated MRI, the forward model of parallel $k$-space data acquisition can be formulated as
\begin{equation}\label{eq:1}\y=\mathcal{M}{\widehat{\x}}\end{equation}
$\widehat{\x},\y\in\mathbb{C}^{N\times N_c}$, $N_c\geq1$ denotes the number of channels, $\widehat{\x}$ is the full-sampled $k$-space data, $i$th column of which denotes the data acquired by the $i$th coil, $\y$ is the under-sampled measurement and $\mathcal{M}$ denotes the sampling pattern. Particularly,
$\mathcal{M}\widehat{\x} = [\mathcal{P}_{\Omega}(\widehat{\x}_1),\ldots,\mathcal{P}_{\Omega}(\widehat{\x}_{N_c})]$, where $\widehat{\x}_i$ denotes $i$th column of $\widehat{\x}$ and $\mathcal{P}_{\Omega}$ denotes the sampling operator on a subset of indices $\Omega\subseteq[N]$. Our task is to interpolate the missing values of $\y$ as accurately as possible. In practice, because of the hardware limitation of MR system, the measurement data is usually mixed with noisy interference, i.e.,
\begin{equation}\label{eq:2}\y^{\delta}=\y+\n\end{equation}
where $\n\in\mathbb{C}^{N\times N_c}$ denotes the noise and $\delta$ denotes the noisy intensity, i.e., $ \delta:=\|\n\|$. Then, the interpolation methods are also required to be robust against noisy interferences. For this purpose, regularization is essential.

\subsection{Zeroth-Order UDN for k-Space PI Model}
As discussed above, due to the fact that there is no need to estimate coil sensitivities, $k$-space PI model has received a lot of attention from industry, among of which SPIRiT is the most prominent example. SPIRIT considers that every point in the grid can be linear predicted by its entire neighborhood in all coils. Given this assumption, any $k$-space data $\widehat{\x}_{i}$ at the $i$th coil can be represented by data from other coils with kernel $\w_{i,n}$. Then, the $k$-space PI regularization model reads:
\begin{equation}\label{slr:1}\left\{\begin{aligned}
		\min_{\widehat{\x}\in\mathbb{C}^{N\times N_c}} &R(\widehat{\x}):=\sum_{i=1}^{N_c}\left\|\widehat{\x}_i-\sum_{n=1}^{N_c}\widehat{\x}_n\otimes{\w_{i,n}}\right\|\\
		\text{s.t.}~~&\mathcal{ M}\widehat{\x}=\y.
	\end{aligned}\right.\end{equation}
where $R$ is the so called self-consistency regularizer.
Algorithmically, the POCS is an effective zeroth-order algorithm for solving problem (\ref{slr:1}), which carries out the following updates:
\begin{equation*}\left\{\begin{aligned}
		\widehat{\x}_i^{k+\frac{1}{2}}&=\sum_{n=1}^{N_c}\widehat{\x}_n^k\otimes{\w_{i,n}}\\
		\widehat{\x}^{k+1}&=\mathcal{P}_{\mathcal{C}}(\widehat{\x}^{k+\frac{1}{2}})
	\end{aligned}\right.\end{equation*}
where $\widehat{\x}^{k}=[\widehat{\x}_1^{k},\ldots,\widehat{\x}_{N_c}^{k}]$ and $\mathcal{C}=\{\widehat{\x}\in\mathbb{C}^d|\mathcal{M}\widehat{\x}=\y\}$.
Looking closely at the above iterates, we can see that the POCS algorithm only calls the zeroth-order information of regularizer $R$ and does not call higher-order information.
Leveraging the idea of UDN, we release the linear convolution kernel (1-convolutional layer) in $R$ as a learnable multi-layer CNN module $\Phi_{\phi}$ with parameters $\phi$ and
learns it in an end-to-end fashion. Particularly, the recursion of the unrolled POCS algorithm reads:
\begin{equation}\label{gpocs}\left\{\begin{aligned}
		\widehat{\x}^{k+\frac{1}{2}}&=\Phi_{\phi}({\x}^k)\\
		\widehat{\x}^{k+1}&=\mathcal{P}_{\mathcal{C}}(\widehat{\x}^{k+\frac{1}{2}}).
	\end{aligned}\right.\end{equation}
Unlike the traditional POCS algorithm, which learns the weights $\{\w_{i,n}\}$ only from a small calibration region of measurement itself, unrolled POCS (\ref{gpocs}) pre-learns the parameters $\phi$ from other full-sampled training data. Therefore, this scheme can enable calibration-free parallel MRI.
Specifically, the unrolled POCS (\ref{gpocs}) can be summed up as a zeroth-order solution algorithm for the following generalized PI regularization model:
\begin{equation}\label{slr:1:1}\left\{\begin{aligned}
		\min_{\widehat{\x}\in\mathbb{C}^{N\times N_c}} &R(\widehat{\x}):=\|\widehat{\x}-\Phi_{\phi}(\widehat{\x})\|^2_F\\
		\text{s.t.}~~&\mathcal{ M}\widehat{\x}=\y.
	\end{aligned}\right.\end{equation}
That is to say, the output of (\ref{gpocs}) can inherit the explainable and predictable nature of model (\ref{slr:1:1}) completely. (\ref{slr:1:1}) is a generalization of SPIRiT model (\ref{slr:1}),
of which the linear predictable prior is extended to nonlinear predictability. If $\Phi_{\phi}$ is chosen as a one convolutional layer, (\ref{slr:1:1}) reduces to (\ref{slr:1}) absolutely.

\begin{figure}[thbp]
	\begin{center}
		\subfigure[]{\includegraphics[width=0.8\textwidth,height=0.25\textwidth]{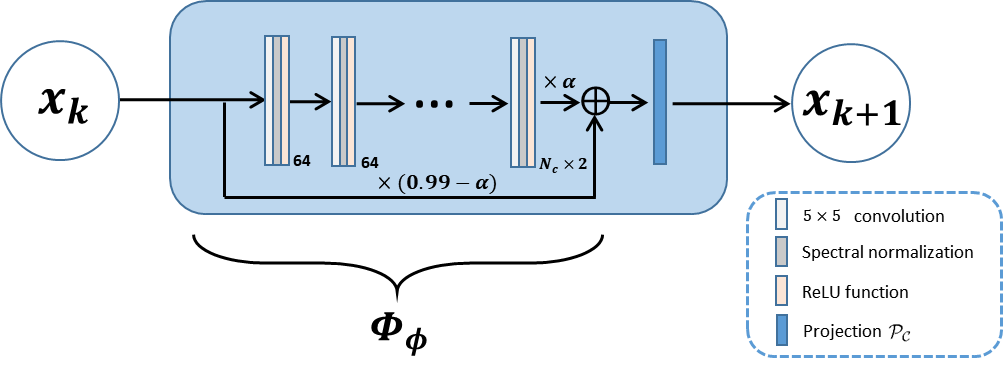}}\\
		\subfigure[]{\includegraphics[width=0.8\textwidth,height=0.3\textwidth]{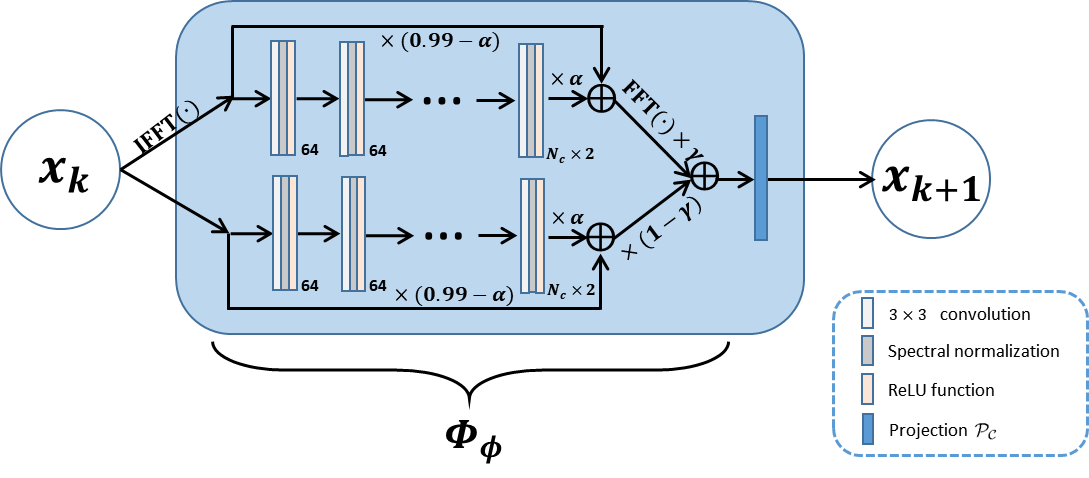}}
	\end{center}
	\caption{The schematic diagram of network architecture of unrolled POCS algorithm: (a) $k$-space unrolled POCS: the self-consistency $\Phi_{\phi}$ is generalized by a five-layer nonexpansive residual CNN in $k$-space, and $\mathcal{P}_{\mathcal{C}}$ adopts the general orthogonal projection; (b) $k$-space and image domain hybrid unrolled POCS: the self-consistency $\Phi_{\phi}$ is linearly composed of two five-layer nonexpansive residual CNNs in $k$-space and image domain, respectively.}
	\label{f:1}
\end{figure}

The network architecture of $\Phi_{\phi}$ is depicted in Figure \ref{f:1} (a). In order to avoid gradient disappearance, we adopt a residual architecture. Different from the general residual network, we use the paradigm of $(0.99-\alpha)\cdot a +\alpha\cdot b $, $(0\leq\alpha\leq0.99)$ to ensure the nonexpensive property, which is crucial for its convergence. What's more, previous works \cite{eo2018kiki,8434321} have showed that the MR images also exhibit self redundancies in image domain.
we also designed another network architecture depicted in the subfigure (b). The nonexpensive residual architecture above exploits the image domain redundancies and the residual architecture below exploits the $k$-space self-consistency, complementary. The additional image domain residual architecture does not affect the theoretical properties
of the unrolled POCS (\ref{gpocs}).

\subsection{Equilibrated Zeroth-Order UDN for k-Space PI Model}
In general UDN, due to memory limitations, the number of iterations (layers) is usually fixed to some small positive integer. However, it has been found that the deeper the network
layers, the stronger the network expressibility. Then, one attempts to execute a trained shallow UDN recursively to achieve deeper layers. Unfortunately, it usually fails to convergence \cite{Gilton2021Deep}. How to choose an appropriate number of layers is still a challenge in UDN. Recent works on DEQ showed that this challenge can be modeled using a fixed point equation. Following this, we
propose to train and test unrolled POCS algorithm under the framework of DEQ. Firstly, before backpropagating, we execute the unrolled POCS algorithm (\ref{gpocs}) to converge to a fixed point, i.e., $\mathring{\widehat{\x}}=\mathcal{P}_{\mathcal{C}}\left(\Phi_{\phi}(\mathring{\widehat{\x}})\right)$. To make the convergent fixed point $\mathring{\widehat{\x}}$ approximate the true full-sampled $k$-space data $\widehat{\x}$ as tight as possible, we minimize the loss between $\mathring{\widehat{\x}}$ and $\widehat{\x}$, i.e., $\ell(\mathring{\widehat{\x}},{\widehat{\x}})$. According to the chain rule, the calculation of the partial derivative of $\ell(\mathring{\widehat{\x}},{\widehat{\x}})$ at $\phi$ reads:
$$ \frac{\partial\ell(\mathring{\widehat{\x}},{\widehat{\x}})}{\partial\phi}= \frac{\partial\mathcal{P}_{\mathcal{C}}(\Phi_{\phi}(\mathring{\widehat{\x}}))^T}{\partial\phi }
\left(\mathcal{I}-\left.\frac{\partial\mathcal{P}_{\mathcal{C}}(\Phi_{\phi}(\widehat{\x}))}{\partial\widehat{\x} }\right|_{\widehat{\x}= \mathring{\widehat{\x}}} \right)^{-T}  \frac{\partial\ell}{\partial \mathring{\widehat{\x}}}.$$
From above formula, it is easy to verify that the backpropagation for $\phi$ can be calculated on the fixed point $\mathring{\widehat{\x}}$ directly, regardless of how many iterations are carried out, so that the memory does not increase even if the number of layers increases to infinity.

The training process for the DEQ-POCS is depicted in Algorithm \ref{alg:1}. Suppose that the training data is sampled from a certain distribution $\pi_{\widehat{\x}\times\y}$, and the network parameter $\phi_0$ is initialized with normal distribution, $\widehat{\x}_0$ is initialized as $\y$. The algorithms is executed by $K$ epoches. For each epoch, we randomly shuffled the order of the data and then executed the algorithm over the dataset ergodicly. At each iteration, we carry out the unrolled POCS algorithm (\ref{gpocs}) to find a fixed point firstly, and then update the network parameters $\phi$ by a certain optimizer with respect to the loss function $\ell(\mathring{\widehat{\x}}^m,\widehat{\x}^m)$. When the training iteration is complete, the algorithm outputs the self-consistency term $\Phi_{\phi_{KM}}$.
Note that, to find the fixed point faster, we can use the Anderson algorithm \cite{Anderson2011Walker} to accelerate the unrolled POCS algorithm (\ref{gpocs}).
\begin{algorithm}[htb]
	\caption{Training DEQ-POCS Network.}
	\label{alg:1}
	\begin{algorithmic}[1]
		\STATE {\bfseries Input:} training samples $\{(\widehat{\x}^m,\y^m)\}_{m=1}^{M}\sim\pi_{\widehat{\x}\times\y}$;\\
		\STATE {\bfseries Initialize:} $\phi_0$, $\widehat{\x}_0$;\\
		\FOR{$k=0,1,\ldots,K$}
		\STATE $n=0$;
		\FOR{each randomly sampled $(\widehat{\x}^m,\y^m)$, $m\in[M]$ }
		\STATE Carry out (\ref{gpocs}) to find a fixed point:
		$$\mathring{\widehat{\x}}^m=\mathcal{P}_{\{\widehat{\x}|\mathcal{M}\widehat{\x}=\y^m\}}\left(\Phi_{\phi_{kM+n}}(\mathring{\widehat{\x}}^m)\right);$$
		\STATE $\phi_{kM+n+1}=\text{Optimizer}(\ell(\mathring{\widehat{\x}}^m,\widehat{\x}^m);\phi_{kM+n})$;\
		\STATE $n=n+1$;
		\ENDFOR
		\ENDFOR
		\STATE {\bfseries Output:} $\Phi_{\phi_{KM}}.$
	\end{algorithmic}
\end{algorithm}

The testing process for the DEQ-POCS is depicted in Algorithm \ref{alg:2}. Insert the trained $\Phi_{\phi_{KM}}$ into the unrolled POCS algorithm (\ref{gpocs}), and execute it to converge to a fixed point, which is act as the algorithm output.

\begin{algorithm}[htb]
	\caption{Testing DEQ-POCS Network.}
	\label{alg:2}
	\begin{algorithmic}[1]
		\STATE {\bfseries Input:} testing samples $\{\y^n\}_{n=1}^{N}\sim\pi_{\y}$, $\Phi_{\phi_{KM}}$;\\
		\STATE {\bfseries Initialize:} $\widehat{\x}_0$;\\
		\FOR{$n=1,2,\ldots,N$}
		\STATE Carry out (\ref{gpocs}) to find a fixed point:
		$$\mathring{\widehat{\x}}^n=\mathcal{P}_{\{\widehat{\x}|\mathcal{M}\widehat{\x}=\y^n\}}\left(\Phi_{\phi_{KM}}(\mathring{\widehat{\x}}^n)\right);$$
		\ENDFOR
		\STATE {\bfseries Output:} $\{\mathring{\widehat{\x}}^n\}_{n=1}^{N}$.
	\end{algorithmic}
\end{algorithm}

\section{Theoretical Results}\label{sect4}
In this section, firstly, we will show the convergence analysis for the proposed DEQ-POCS network in case that the measurement data is noise-free. Then, we prove that the proposed DEQ-POCS is robust against noisy interference in case that the measurement data contains noise.
\subsection{Convergence}
Before proving our main result, we suppose that the learned self-consistency terms $\Phi_{\phi_{i}}$, $i\in[KM]$ in Algorithm \ref{alg:1} satisfy the following assumption:
\begin{assump}\label{assup:1}
	The learned self-consistency term $\Phi_{\phi_{i}}$, $i\in[KM]$ in Algorithm \ref{alg:1} are $L$-Lipschitz continuous with $0<L<1$.
\end{assump}
The learned self-consistency term $\Phi_{\phi_{i}}$ depicted in Figure \ref{f:1} meets Assumption \ref{assup:1}. Based on it, we have the following result:
\begin{thm}\label{thm:1} Suppose Assumption \ref{assup:1} holds. The unrolled POCS (\ref{gpocs}) in Algorithm \ref{alg:1} and \ref{alg:2} converges to a fixed point globally.
\end{thm}

The proof is presented in supplementary material.
\begin{rmk}
	Although some literatures given the convergence proof of UDN under some conditions, such as,
	Kurdyka-{\L}ojasiewicz condition \cite{9481093,Huang2021Deep}, asymptotically nonexpansive condition \cite{9152164}, uniform decrease condition \cite{8727950}, e.t.c.,
	firstly, these conditions are difficult to verify in practice; secondly, the corresponding theory requires that the number of iterations tends to be infinite and UDN usually unrolls only a few layers.
\end{rmk}

Theorem \ref{thm:1} shows the convergence of DEQ-POCS. Next, we consider how tightly the convergent fixed point approximates the true solution (full-sampled $k$-space data). Before giving the result, we assume the output of Algorithm \ref{alg:1}, i.e., $\Phi_{\phi_{KM}}$, meets the following condition:
\begin{assump}\label{assup:2}
	Let $\mathring{\widehat{\x}}^m$ denote the fixed point of unrolled POCS iterates (\ref{gpocs}) with measurement $\y^m$ and self-consistency $\Phi_{\phi_{KM}}$. If the loss function $\ell$ takes the Frobenius-norm in Algorithm \ref{alg:1}, there exists a constant $\epsilon$ such that $\frac{1}{M}\sum_{m=1}^M\|\mathring{\widehat{\x}}^m-{\widehat{\x}}^m\|_F\leq\epsilon$.
\end{assump}
\begin{rmk}
	Although the loss function is generally nonconvex and nonsmooth with respect to network parameters, there are some literatures \cite{LeCun2015Deep,He_2016_CVPR} empirically indicate that the value of loss usually attenuates to close to 0 as the training goes on, when the depth of network is deeply enough. In theory, there also exit some results to show that, under certain conditions, even the simple stochastic gradient descent algorithm can find global minimizers (0-value loss points) for network training problem \cite{pmlr-v97-allen-zhu19a}.
\end{rmk}

Based on Assumption \ref{assup:1} and \ref{assup:2}, we have the following result:
\begin{thm}\label{thm:2}Suppose Assumption \ref{assup:1} and \ref{assup:2} hold. Then, there exists a constant $B>0$ such that, for any $n\in[N]$, the convergent solution $\mathring{\widehat{\x}}^n$ of Algorithm \ref{alg:2} satisfies:
	\begin{equation}\label{noisefree}\|\mathring{\widehat{\x}}^n-\widehat{\x}^{n}\|_F\leq \left(1+\frac{1}{\sqrt{M}}\right)\lambda+\epsilon\end{equation}
	with probability at least $1-4\exp\left(\frac{-\lambda^2}{2B^2}\right)$, where $\{\widehat{\x}^n\}$ denote the true full-sampled $k$-space data.
\end{thm}

The proof is presented in supplementary material.
To the best of our knowledge, it is the first time to show the reconstruction accuracy guarantee for DL methods in accelerated MRI.
\subsection{Regularity}
In practice, due to the hardware limitation of MR system, the measurement data is usually mixed with noisy interference. For the ill-posedness of accelerated MR reconstruction problem, a small noisy disturbance in the measurement may seriously interfere with the interpolation accuracy. Thus, the robustness (regularity) against noisy interference is very important in designing MR reconstruction methods. In this paper, the proposed DEQ-POCS has the following property:
\begin{thm}\label{thm:3}Suppose Assumption \ref{assup:1} and \ref{assup:2} hold. When the measurement data contains noise, i.e., $\y^{n,\delta}=\y^n+\n^n$ for any $n\in[N]$ with noisy level $ \delta=\|\n^n\|_F$, the convergent solution $\{\mathring{\widehat{\x}}^{\delta,n}\}$ of Algorithm \ref{alg:2} satisfies:
	$$\|\mathring{\widehat{\x}}^{\delta,n}-\widehat{\x}^{n}\|_F\leq\frac{\delta}{1-L}+\left(1+\frac{1}{\sqrt{M}}\right)\lambda+\epsilon$$
	with probability at least $1-4\exp\left(\frac{-\lambda^2}{2B^2}\right)$, where $\{\widehat{\x}^n\}$ denote the true full-sampled $k$-space data.
\end{thm}
The proof is presented in supplementary material. From above theorem, we can see that a $\delta$-level noisy interference in measurement at most causes $\delta/(1-L)$-level error in interpolated result $\mathring{\widehat{\x}}^{\delta,n}$ compared to noisy-free case, and the error will disappear as $\delta\rightarrow0$. In other words, the interpolated result $\mathring{\widehat{\x}}^{\delta,n}$ is robust against noisy measurement $\y^{n,\delta}$.
\section{Implementation}\label{sect5}
The evaluation were performed on two multichannel $k$-space data with various $k$-space trajectories. The details of the $k$-space data are as follows:
\subsection{Data Acquisition}
\subsubsection{Knee data}
Firstly, we tested our proposed method on a knee MRI data \footnote{\url{http://mridata.org/}}. The raw data was acquired from a 3T Siemens scanner. The number of coils was 15 and the 2D Cartesian turbo spin echo (TSE) protocol was used. The parameters for data acquisition are as follows: the repetition time (TR) was 2800ms, the echo time (TE) was 22ms, the matrix size was $768\times 770\times 1$ and the field of view (FOV) was $280 \times 280.7 \times 4.5 \text{mm}^3$. Particularly, the readout oversampling was removed by transforming the $k$-space to image, and cropping the center $384 \times 384$ region. Fully sampled multichannel knee images of nine volunteers were collected out of which data from seven subjects (included 227 slices) were used for training, while
the data from the rest two subjects (included 64 slices) were used for testing.

\subsubsection{Human brain data}
To verify the generalization of the proposed method, we also tested it on a human brain MRI data \footnote{\url{https://drive.google.com/file/d/1qp-l9kJbRfQU1W5wCjOQZi7I3T6jwA37/view?usp=sharing}}, which was collected by \cite{8434321}. This MRI data were acquired using a
3D T2 fast spin echo with an extended echo train acquisition (CUBE) sequence with Cartesian readouts using a 12-channel head coil. The matrix dimensions were $256\times232\times208$ with 1 mm isotropic resolution.
The training data contains 360 slices $k$-space data from four subjects and the testing data contains 164 slices $k$-space data from two subjects. Each slice is of spatial dimension $256\times232$.

\subsubsection{Sampling patterns}
\begin{figure}[thbp]
	\begin{center}
		\subfigure{\includegraphics[width=0.8\textwidth,height=0.25\textwidth]{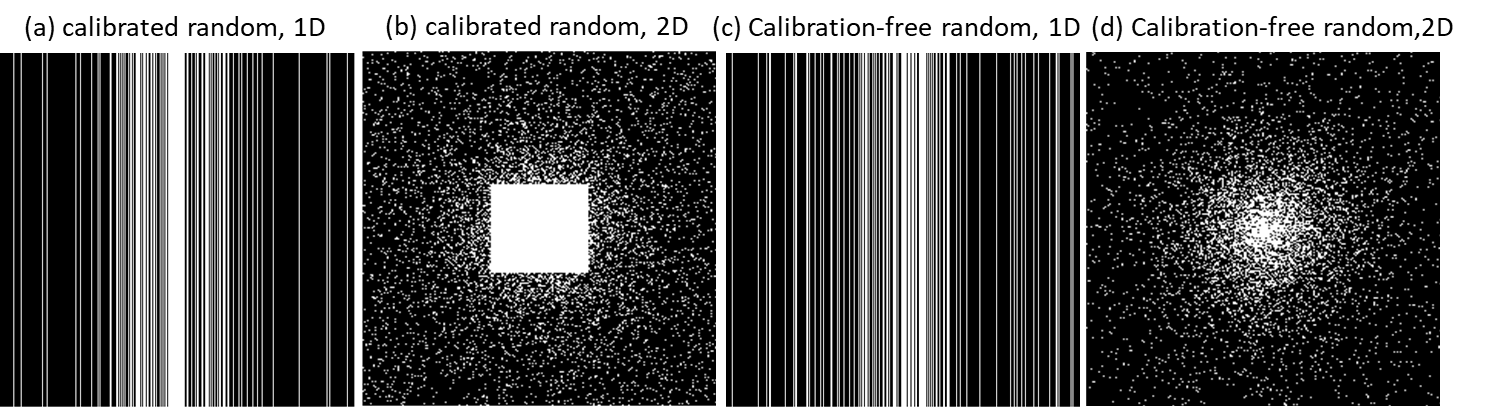}}
	\end{center}
	\caption{Various sampling patterns: (a) 1-D calibrated random under-sampling at $R=4$, (b) 2-D calibrated random under-sampling at $R=6$, (c) 1-D calibration-free random under-sampling at $R=4$, and (d) 2-D calibration-free random under-sampling at $R=10$.}
	\label{f2}
\end{figure}

Four different types of under-sampling patterns were tested, i.e., 1-D and 2-D calibrated and calibration-free radom trajectories. For the calibrated trajectories, $64\times64$  auto-calibration signal (ACS)  regions and 16 ACS lines were fully sampled, respectively. A visualization of these sampling patterns is depicted in Figure \ref{f2}.
\subsection{Network Architecture}
The schematic diagram of self-consistence module $\Phi_{\phi}$ architecture is illustrated in Figure \ref{f:1}. Experientially, it has been found that the higher the degree of network parameterization, the stronger the expression ability. Then, we expand $\Phi_{\phi}$ into the following ten modules:
$$\Phi_{\phi}:=[\Phi^1_{\phi},\ldots,\Phi^{10}_{\phi}]$$
and generalize each module to a deep neural network with architecture depicted in Figure \ref{f:1}. To find the fixed point faster, we use the Anderson algorithm to accelerate the unrolled POCS algorithm (\ref{gpocs}) whose code is available on this page \footnote{\url{http://implicit-layers-tutorial.org/}}. Particularly, at each iteration in Algorithm \ref{alg:1} and \ref{alg:2}, we carry out the Anderson accelerated algorithm with five iteration.

\subsection{Network Training}
The optimizer in Algorithm \ref{alg:1} is chosen the ADAM \cite{kingma2014adam} optimizer with $\beta_1=0.9, \beta_2=0.999$ with respect to the $\ell_2$-norm loss in $k$-space. The size of mini batch is 1 and the number of epochs is 500. The learning rate is set as $10^{-4 }$.
The labels for the network were the fully sampled $k$-space data. The input data for the network was the regridded down-sampled $k$-space data from 1-D and 2-D random trajectories. The details of the downsampling procedure has be discussed on the above. Without specific instructions, we train the network separately for different trajectories.
The models were implemented on an Ubuntu 16.04 LTS (64-bit) operating system equipped with an Intel Xeon Gold5120 central processing unit (CPU) and Tesla V100 graphics processing unit (GPU, 32 GB memory) in the open framework PyTorch 1.7.1 \cite{paszke2019pytorch} with CUDA and CUDNN support.
\subsection{Performance Evaluation}
In this paper, the quantitative evaluation were all calculated on image domain. The image is derived by the inverse Fourier transform followed by an element-wise square-root of sum-of-the squares (SSoS) operation, i.e. $\z[n]=(\sum_{i=1}^{N_c}|\x_i[n]|^2)^{\frac{1}{2}}$, where $\z[n]$ denotes the $n$-th element of image $\z$, and $\x_i[n]$ denotes the $n$-th element of the $i$th coil image $\x_i$. For quantitative evaluation, the peak signal-to-noise ratio (PSNR), normalized mean square error (NMSE) value and structural similarity (SSIM) index \cite{1284395} were adopted.

\section{Experimental Results }\label{sect6}
\subsection{Comparative Studies}
In this section, we test our proposed DEQ-POCS with $k$-space and hybrid architectures depicted in Figure \ref{f:1}, dubbed K-DEQ-POCS and H-DEQ-POCS for short, on the knee and brain datasets. To demonstrate the effectiveness of our methods, a series of extensive comparative experiments were studied. In particular, we compared with the following $k$-space interpolation networks or algorithms: SPIRiT-POCS \cite{Lustig2010SPIRiT}, Deep-SLR \cite{9159672} based on $k$-space and hybrid architectures, dubbed K-Deep-SLR and H-Deep-SLR.
For the sake of fairness, K-Deep-SLR and H-Deep-SLR are unrolled to 10 layers to share the same number of parameters as K-DEQ-POCS and H-DEQ-POCS respectively, for which we develop a PyTorch-based implementation based on their publicly available TensorFlow codes\footnote{\url{https://github.com/anikpram/Deep-SLR}}.

\subsection{Experiments Without Additional Noisy Interference}
\begin{figure}[thbp]
	\begin{center}
		\subfigure{\includegraphics[width=0.95\textwidth,height=0.48\textwidth]{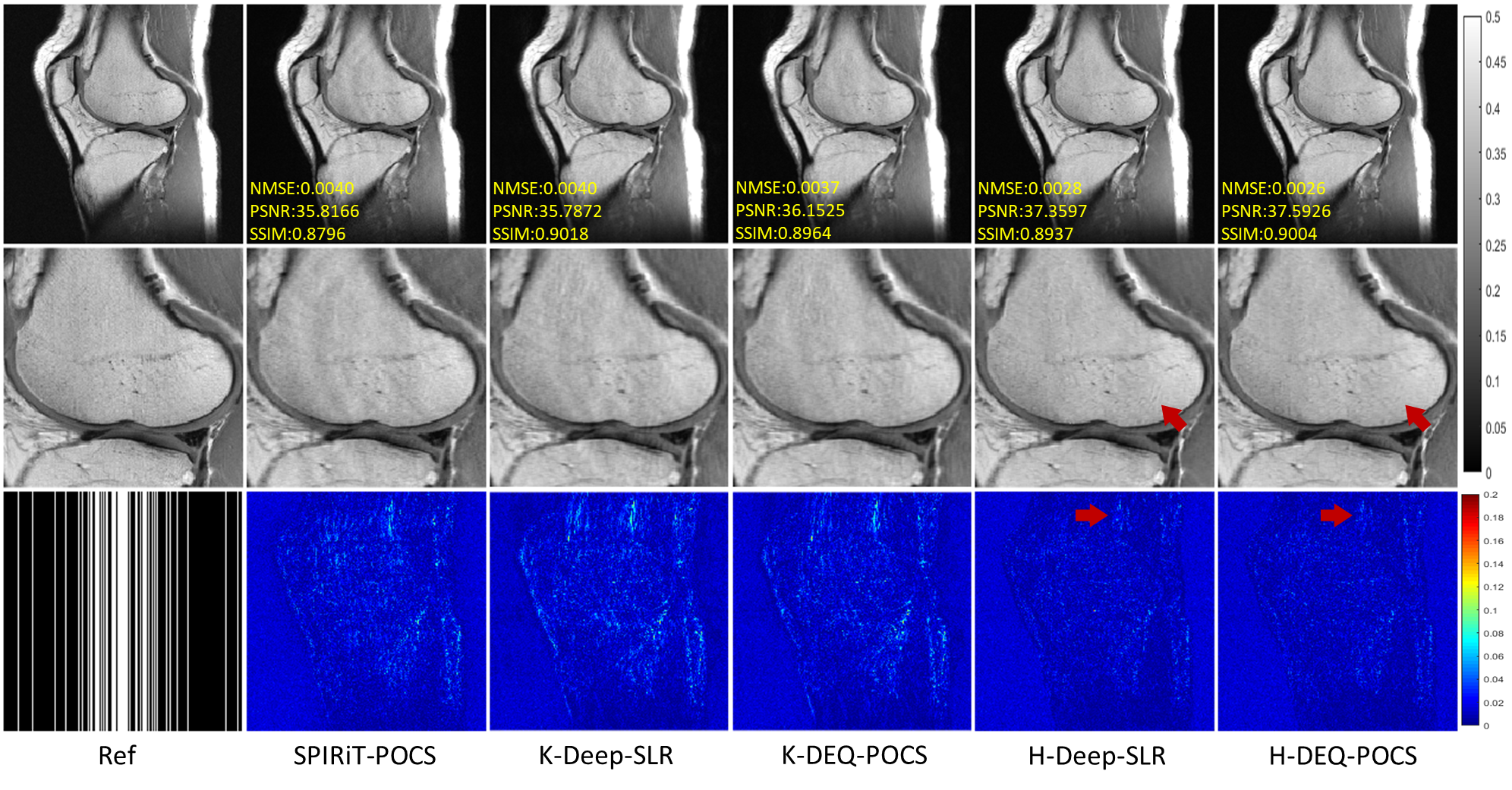}}
	\end{center}
	\caption{Reconstruction results under 1-D calibrated random under-sampling at $R=4$. The values in the corner are NMSE/PSNR/SSIM values of each slice. Second and third rows illustrate the enlarged and error views, respectively. The gray scale of the reconstructed images and the color bar of the error images are at the right of the figure.}
	\label{f5}
\end{figure}

In this section, we test our proposed K-DEQ-POCS and H-DEQ-POCS and comparative networks or algorithms without additional noisy interference.
Figure \ref{f5} shows the reconstruction results of the knee data using various methods under 1-D calibrated random trajectory with acceleration factor 4. As shown in Figure \ref{f5}, for the networks (or algorithms) with single $k$-space architecture including K-Deep-SLR, K-DEQ-POCS and SPIRiT-POCS algorithm, the aliasing pattern still remains in the reconstructed images. But, form the enlarged view, we can observe that our K-DEQ-POCS slightly outperforms K-Deep-SLR and SPIRiT-POCS.
For the networks with $k$-space and image domain hybrid architectures, the aliasing pattern can be effectively suppressed.
From the enlarged view of Figure \ref{f5}, it is not difficult to find that the image reconstructed by H-Deep-SLR contains slight artifacts, while our H-DEQ-POCS does not, behind which the reason is mainly attributed to the fact that H-DEQ-POCS carries out more iterations to solve the forward model (\ref{eq:1}) more accurately.

\begin{figure}[thbp]
	\begin{center}
		\subfigure{\includegraphics[width=0.95\textwidth,height=0.35\textwidth]{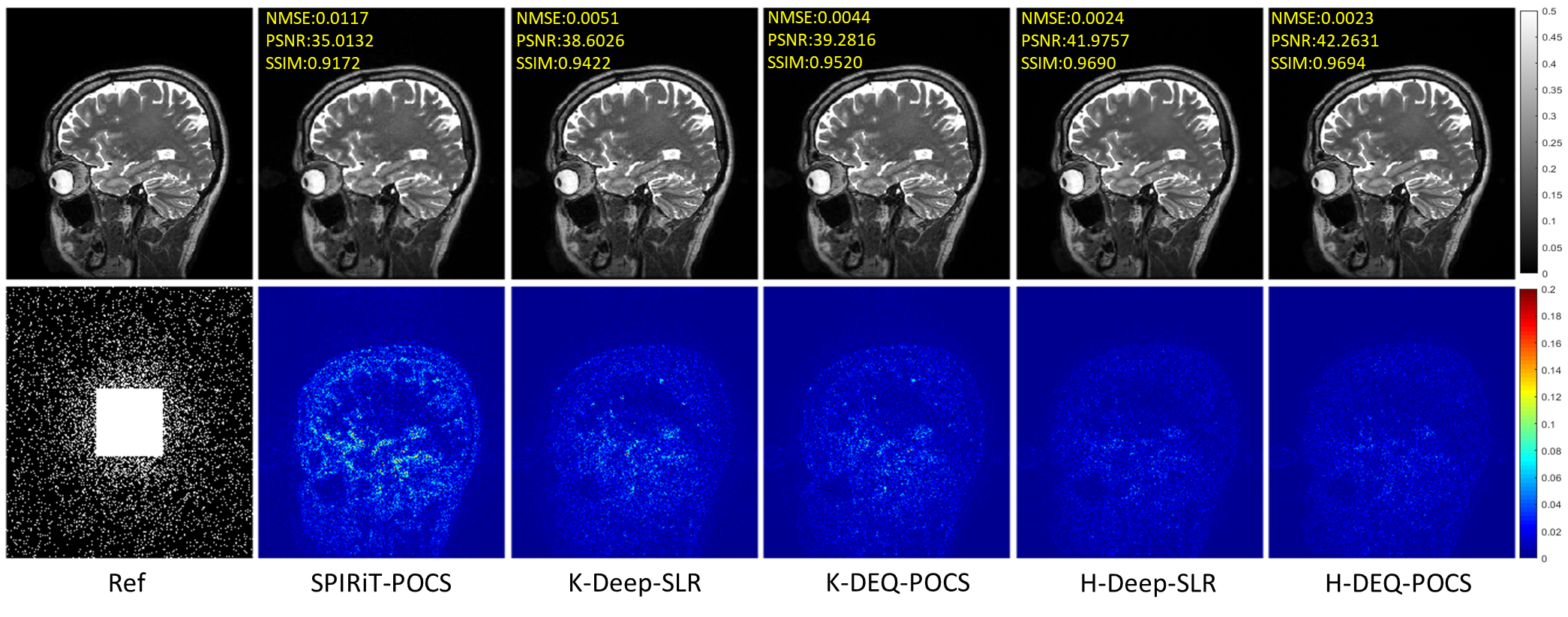}}
	\end{center}
	\caption{Reconstruction results under 2-D calibrated random under-sampling at $R=6$. The values in the corner are NMSE/PSNR/SSIM values of each slice. Third row illustrates the error views. The gray scale of the reconstructed images and the color bar of the error images are at the right of the figure.}
	\label{f4}
\end{figure}

Figure \ref{f4} shows the reconstruction results of the brain data using various methods under the 2-D calibrated random trajectory with acceleration factor 6. For the networks (or algorithms) with single $k$-space architecture, our method K-DEQ-POCS outperforms K-Deep-SLR and SPIRiT-POCS with respect to noise artifacts amplification.
For the networks with hybrid architectures, both H-DEQ-POCS and H-Deep-SLR have achieved good performance. If we take a close look at the error view, we can see that our method has less error in the neat edges.

The competitive quantitative results of above methods are shown in Table \ref{tab:1}. For knee data and brain data, our method consistently outperforms the other comparative methods. Therefore, as characterized by visual and quantitative evaluations, the above experiments confirm the competitiveness of our method under the case with no additional noisy interference.
\begin{table}
	\begin{center}
		\caption{Quantitative comparison for various methods on the knee and brain data under the calibrated trajectories.}\label{tab:1}
		\setlength{\tabcolsep}{2mm}{
			\begin{tabular}{l|l|ccc}
				\hline
				\multicolumn{ 2}{c}{ Datasets} & \multicolumn{ 3}{|c}{Quantitative Evaluation}  \\
				\multicolumn{ 2}{c|}{ \& Methods   } &NMSE &PSNR(dB)&SSIM  \\
				\hline
				\multirow{5}{*}{Knee}
				& SPIRiT-POCS  &0.0041$\pm$0.0013&36.01$\pm$1.72&0.88$\pm$0.02. \\
				\cline{2-5}
				& K-Deep-SLR  &0.0040$\pm$0.0011&36.05$\pm$1.56&0.90$\pm$0.02\\
				\cline{2-5}
				& K-DEQ-POCS &0.0039$\pm$0.0011&36.20$\pm$1.68&0.90$\pm$0.02\\
				\cline{2-5}
				&  H-Deep-SLR &0.0028$\pm$0.0010&37.65$\pm$1.42&0.89$\pm$0.02\\
				\cline{2-5}
				&  H-DEQ-POCS &\textcolor{red}{0.0026$\pm$0.0009}&\textcolor{red}{37.94$\pm$1.42}&0.90$\pm$0.02\\
				\hline
				\multirow{5}{*}{Brain}
				& SPIRiT-POCS  &0.0123$\pm$0.0007&35.07$\pm$0.92&0.92$\pm$0.01 \\
				\cline{2-5}
				& K-Deep-SLR  &0.0051$\pm$0.0020&39.20$\pm$1.51&0.94$\pm$0.02\\
				\cline{2-5}
				& K-DEQ-POCS &0.0045$\pm$0.0015&39.62$\pm$1.26&0.95$\pm$0.01\\
				\cline{2-5}
				&  H-Deep-SLR &0.0024$\pm$0.0006&42.37$\pm$1.00&0.97$\pm$0.01\\
				\cline{2-5}
				& H-DEQ-POCS &\textcolor{red}{0.0023$\pm$0.0005}&\textcolor{red}{42.38$\pm$0.87}&0.97$\pm$0.01\\
				\hline
		\end{tabular}}
	\end{center}
\end{table}

\subsection{Interference on Measurement}
\textcolor{black}{
In practice, the measurement from different MR system often contain different level (Gaussian) noise due to magnetic field inhomogeneity and hardware restriction.
In order to get closer to the actual simulation, in this experiment, we will test various methods when the measurement data contains additional $\delta$ intensity noise, i.e., $\y^{\delta}:=\y+\n$.}

\begin{figure}[thbp]
	\begin{center}
		\subfigure{\includegraphics[width=0.95\textwidth,height=0.48\textwidth]{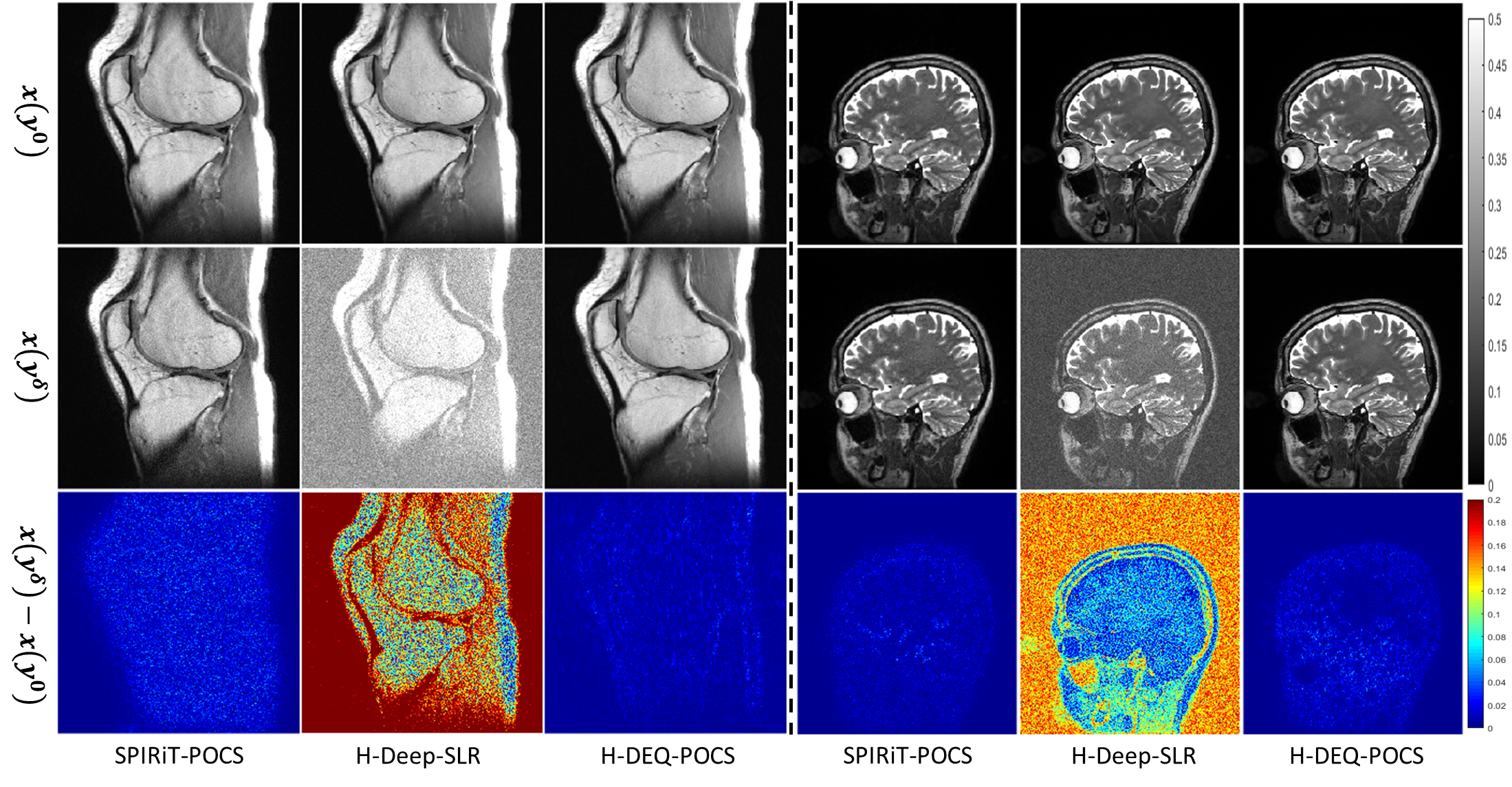}}
	\end{center}
	\caption{First and second rows illustrate the reconstruction results from clean measurement data $\y^{0}:=\y$ and noisy measurement data $\y^{\delta}:=\y+\n$ (which contains 1$\%$ Gaussian noise on knee and 0.5$\%$ Gaussian noise on brain measurement data, respectively). The third raw illustrates the error caused by noisy measurement.}
	\label{f6}
\end{figure}

Figure \ref{f6} shows the reconstruction results on the knee and brain dataset using various methods under the calibrated 4x 1-D and 6x 2-D random trajectories with clean and noisy measurement, respectively.
Obviously, compared to the results from clean measurement, for H-Deep-SLR, the noise artifacts amplified from noisy interference in the measurement result in serious distortions.
For traditional SPIRiT-POCS algorithm and our H-DEQ-POCS, the noise amplification is significantly lower than H-Deep-SLR.
This experimental result verifies well the validity of Theorem \ref{thm:3}.
From this experiment, we conclude that our proposed method can not only achieves the state-of-the-art performance like general UDN, but also is robust against noisy interferences like traditional iterative algorithms.

\subsection{Interference on Initial Input}
Recent work \cite{Antun30088} pointed out that DL methods typically yield unstable reconstruction with different initial input.
In practice, it is difficult to ensure that the initial input in testing is consistent with it in training.
Therefore, in this experiment, we test the robustness of various method against interferences on initial input, i.e., $\widehat{\x}_0^{\delta}:=\widehat{\x}_0+\n$.

\begin{figure}[thbp]
	\begin{center}
		\subfigure{\includegraphics[width=0.95\textwidth,height=0.48\textwidth]{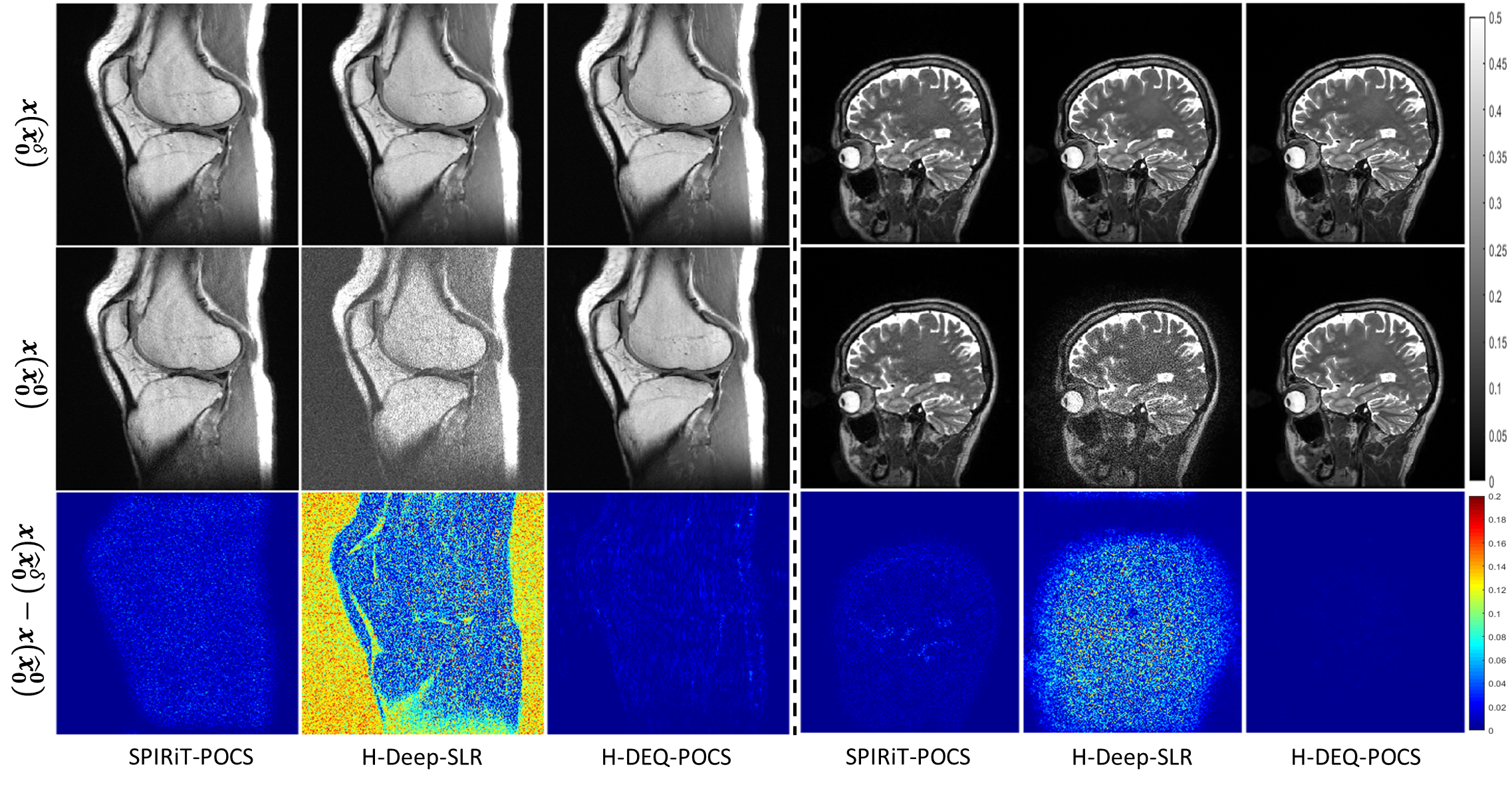}}
	\end{center}
	\caption{First and second rows illustrate the reconstruction results from clean initial input $\widehat{\x}_{0}^{0}:=\widehat{\x}_0$ and noisy initial input $\widehat{\x}_{0}^{\delta}:=\widehat{\x}_0+\n$
	 (which contains 50$\%$ Gaussian noise on initial input for knee and 5$\%$ Gaussian noise on initial input for brain data, respectively). The third raw illustrates the error caused by noisy interference on initial input.}
	\label{f7}
\end{figure}

Figure \ref{f7} shows the reconstruction results on the knee and brain dataset using various methods under the calibrated 4x 1-D and 6x 2-D random trajectories with clean and noisy initial input, respectively.
Obviously, for H-Deep-SLR, the noise artifacts caused by the noisy input $\widehat{\x}_0^{\delta}$ lead to serious distortions.
However, we can observe that the noisy initial input has very little influence on the performances of traditional SPIRiT-POCS algorithm and our H-DEQ-POCS. This experimental result well confirms our previous claim that
the learning process of the proposed DEQ-POCS is only based on convergent fixed points and independent of initial input.

\subsection{Varying on Sampling Pattern}
Next, we test whether our proposed method trained under one sampling trajectory can be migrated to another. Specifically, we will test the trained models in above experiments on
varying 4x 1-D and 10x 2-D calibration-free random trajectories under-sampled measurement. In this experiment, because there is no ACS data, SPIRiT-POCS is no longer suitable.

\begin{figure}[thbp]
	\begin{center}
		\subfigure{\includegraphics[width=0.95\textwidth,height=0.48\textwidth]{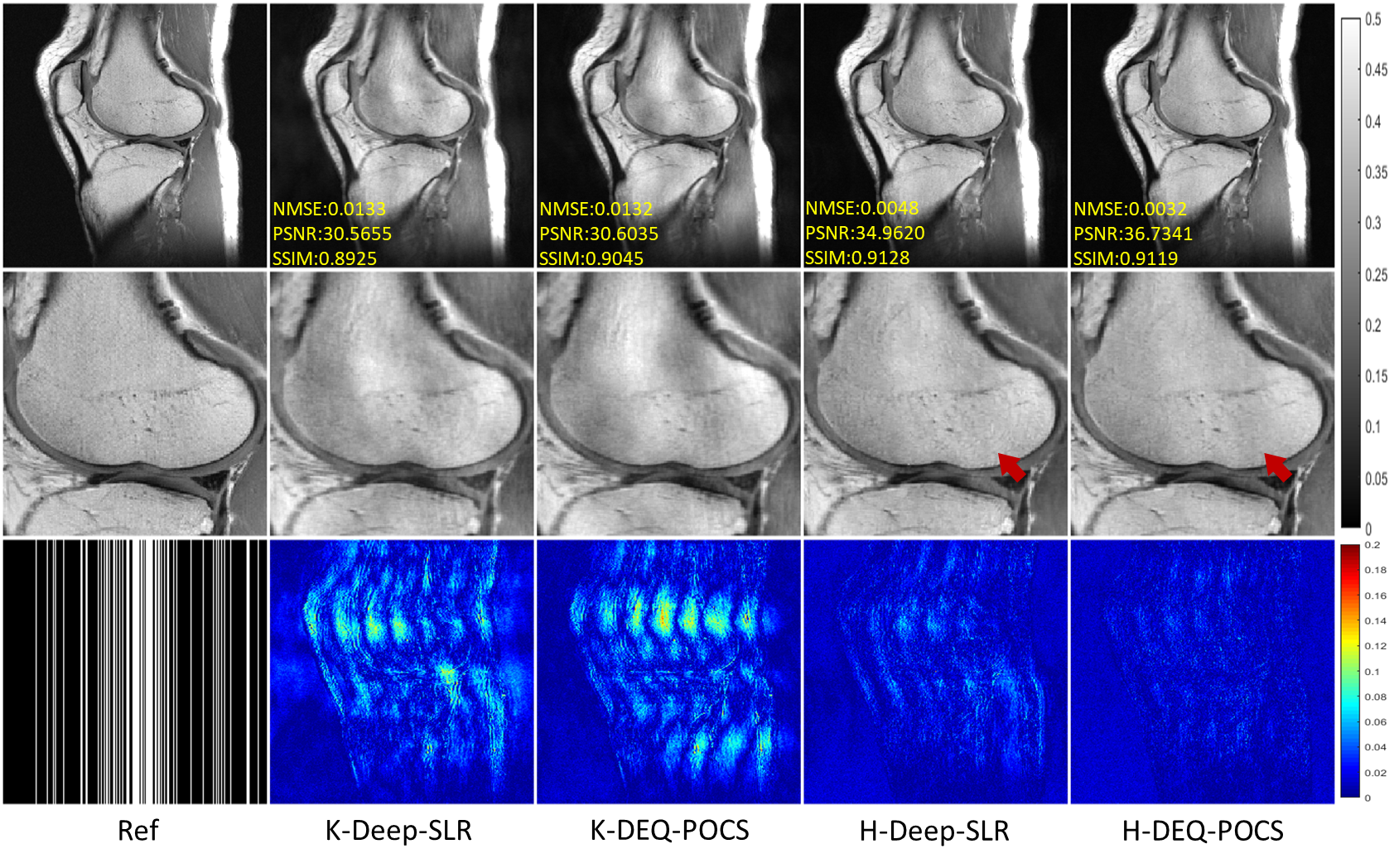}}
	\end{center}
	\caption{Reconstruction results under calibration-free 1-D random under-sampling at $R=4$. The values in the corner are NMSE/PSNR/SSIM values of each slice. Second and third rows illustrate the enlarged and error views, respectively. The gray scale of the reconstructed images and the color bar of the error images are at the right of the figure..}
	\label{f8}
\end{figure}

\begin{figure}[thbp]
	\begin{center}
		\subfigure{\includegraphics[width=0.95\textwidth,height=0.35\textwidth]{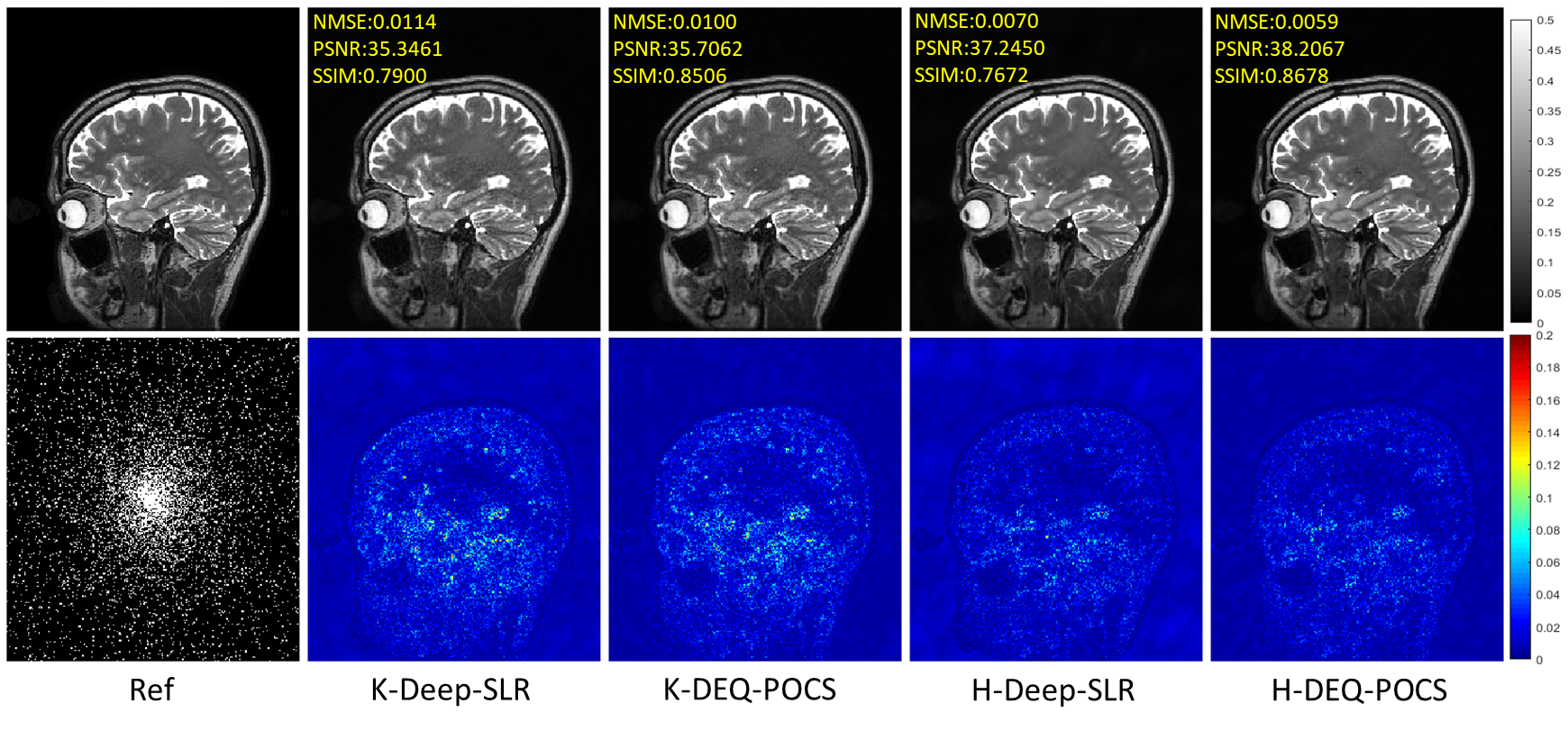}}
	\end{center}
	\caption{Reconstruction results under calibration-free 2-D random under-sampling at $R=10$. The values in the corner are NMSE/PSNR/SSIM values of each slice. Second row illustrates the error views. The gray scale of the reconstructed images and the color bar of the error images are at the right of the figure.}
	\label{f9}
\end{figure}

\begin{table}[thbp]
	\begin{center}
		\caption{Quantitative comparison for various methods on the knee and brain data under the calibration-free trajectories.}\label{tab:2}
		\setlength{\tabcolsep}{2mm}{
			\begin{tabular}{c|c|cccc}
				\hline
				\multicolumn{ 2}{c}{ Datasets} & \multicolumn{ 3}{|c}{Quantitative Evaluation}  \\
				\multicolumn{ 2}{c|}{ \& Methods   } &NMSE &PSNR(dB)&SSIM   \\
				\hline
				\multirow{4}{*}{Knee}
				& K-Deep-SLR  &0.0155$\pm$0.0069&30.32$\pm$2.00&0.88$\pm$0.03\\
				\cline{2-5}
				& K-DEQ-POCS &{0.0136$\pm$0.0039}&{30.71$\pm$2.00}&{0.90$\pm$0.02}\\
				\cline{2-5}
				&  H-Deep-SLR &0.0041$\pm$0.0010&35.92$\pm$1.63&0.91$\pm$0.01\\
				\cline{2-5}
				&  H-DEQ-POCS &\textcolor{red}{0.0030$\pm$0.0009}&\textcolor{red}{37.34$\pm$1.60}&\textcolor{red}{0.92$\pm$0.01}\\
				\hline
				\multirow{4}{*}{Brain}
				& K-DEEP-SLR  &0.0130$\pm$0.0021&34.88$\pm$0.80&0.79$\pm$0.04\\
				\cline{2-5}
				& K-DEQ-POCS &{0.0115$\pm$0.0020}&{35.43$\pm$0.82}&{0.83$\pm$0.04}\\
				\cline{2-5}
				&  H-DEEP-SLR &0.0081$\pm$0.0015&36.98$\pm$0.71&0.73$\pm$0.05\\
				\cline{2-5}
				&  H-DEQ-POCS &\textcolor{red}{0.0066$\pm$0.0012}&\textcolor{red}{37.84$\pm$0.69}&\textcolor{red}{0.85$\pm$0.04}\\
				\hline
		\end{tabular}}
	\end{center}
\end{table}

Figure \ref{f8} shows the reconstruction results of trained models on varying calibration-free 1-D random under-sampling with acceleration factor 4 for knee data. As shown in Figure \ref{f8}, we observe that the networks with single $k$-space architectures are greatly affected by the varying on sampling pattern, but the migration of the networks with hybrid architectures to this calibration-free pattern is relatively successful.
However, if we take a closer look at the enlarged view of Figure \ref{f8}, the aliasing pattern still remains in the result of H-Deep-SLR. In Figure \ref{f4},
H-Deep-SLR and H-DEQ-POCS in Figure \ref{f4} perform almost equally, but with varying on sampling patterns, our H-DEQ-POCS is significantly better than H-Deep-SLR, which means that H-DEQ-POCS is robust in varying sampling patterns. Figure \ref{f9} shows the reconstruction results of trained models on varying calibration-free 2-D random under-sampling with acceleration factor 10 for brain data. The experimental results are consistent with the above experimental performance, which further verifies the robustness of our method in terms of varying sampling patterns.
The competitive quantitative results of above methods are shown in Table \ref{tab:2}, which confirms the competitiveness of our method under the case with varying sampling patterns

\section{Discussion}\label{sect7}
In this paper, we proposed a zeroth-order UDN i,e., generalized POCS algorithm, for $k$-space PI regularization problem in an equilibrated manner, dubbed DEQ-POCS, which is guaranteed to converge to a fixed pint globally.
Since the output of DEQ-POCS can be covered by a generalized $k$-space PI regularization model, DEQ-POCS can inherit its explainable and predictable nature completely.
The experimental results shown in the previous section demonstrated that our proposed method has potential advantages compared to existing $k$-space UDN and traditional methods.
We also prove that DEQ-POCS is robust against noisy interference. Accordingly, the above experimental results showed that proposed method shows stable performance to deal with interference in measurement, interference in initial input and varying on sampling pattern, which confirms our theoretical claim well. However, there are still several points worth improving for our method.

In theory, Theorem \ref{thm:2} showed that the convergent fixed point of DEQ-POCS enjoys a tight complexity guarantee to approximate the true MR image (full-sampled $k$-space data). However,
this result cannot guarantee that DEQ-POCS can reconstruct the true MR image completely.
Inspired by CS or matrix completion theory, leveraging the complete recovery conditions, such as null space property (NSP) and restricted isometry property (RIP), e.t.c., to driven DL based MR reconstruction methods with completely recovery property will be our future work.

In Theorem \ref{thm:3}, we have proved DEQ-POCS is an iterative regularization method.
In case the measurement data contains noise, for inverse problem, some literatures \cite{jiao2016alternating} have illustrated that terative methods always show asemiconvergence property, i.e., the iterate becomes close to the sought solution at the beginning of iteration; however, after a critical number of iterations, the iterate leaves the sought solution far away as the iteration proceeds.
Therefore, a suitable terminated criterion is essential for iterative regularization methods, but, which is difficult to implement without knowing the noisy intensity beforehand.
Recent works show that the Bayesian approach can avoid the need for early termination and additionally provide reconstruction uncertainty \cite{Cheng_2019_CVPR,Laves2020Uncertainty}. Redesigning our DEQ-POCS
with Bayesian framework may prove to be a fruitful direction for future work.


\section{Conclusion}\label{sect8}
In this paper, we proposed a zeroth-order UDN i,e., generalized POCS algorithm, for $k$-space PI regularization problem in an equilibrated manner, dubbed DEQ-POCS.
Different from first-order UDN, DEQ-POCS can inherit explainable and predictable nature $k$-space of a generalized PI regularization model completely.
Theoretically, we proved that the DEQ-POCS is guaranteed to converge to a fixed point which enjoys a tight complexity to approximate the true full-sampled $k$-space data. Furthermore, in terms of robustness, we proved that DEQ-POCS is robust against noisy interferences and sampling pattern varying.
Experimentally, we verified that the proposed DEQ-POCS outperforms existing $k$-space UDN and traditional methods. We believe that our method can be a powerful framework for accelerated MRI and its further development of this kind of methods may enable even bigger gains in the future.

\section{Appendix}

\subsection{Additional Experiments}
\begin{figure}[thbp]
	\begin{center}
		\subfigure{\includegraphics[width=0.95\textwidth,height=0.48\textwidth]{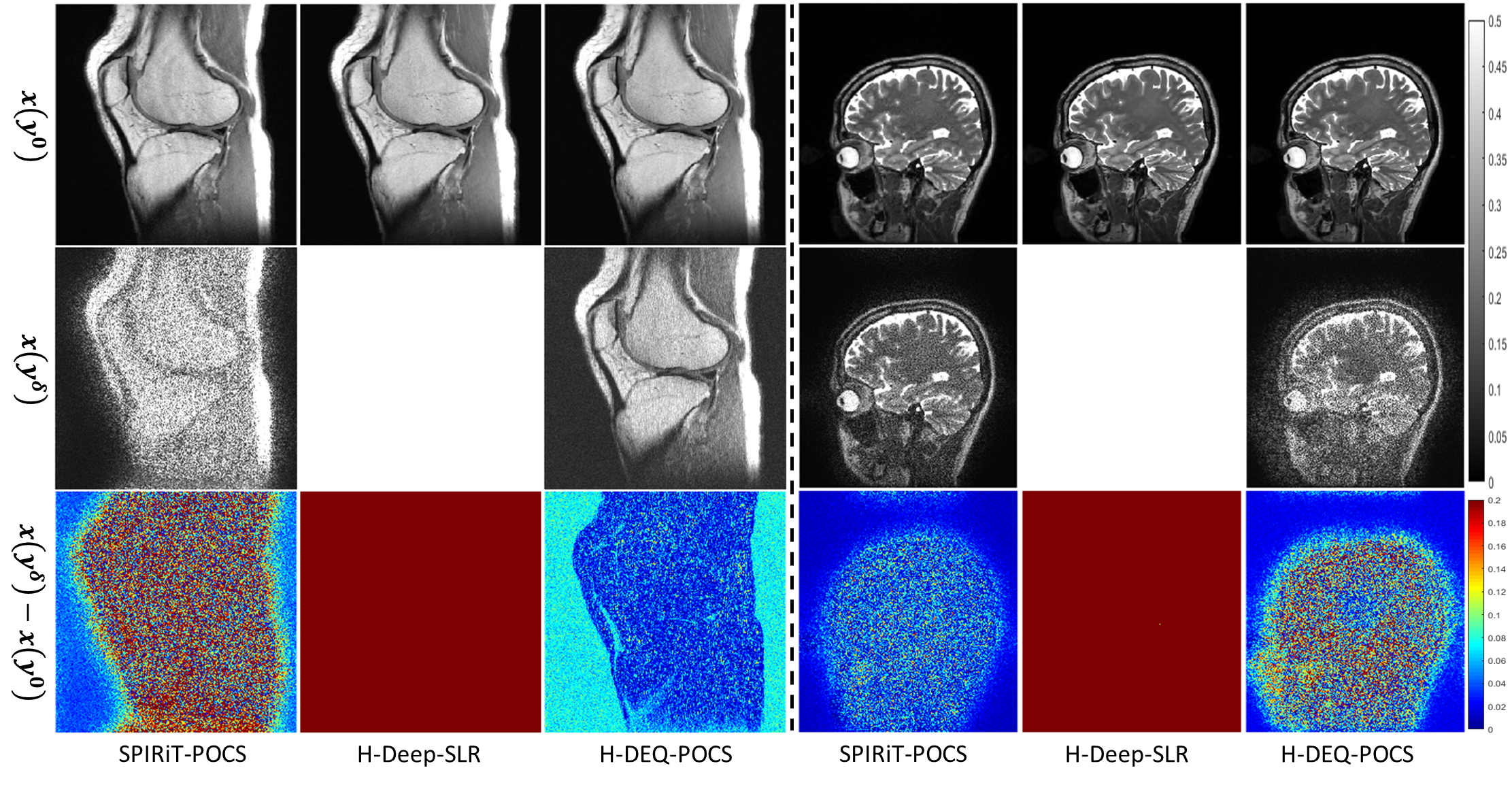}}
	\end{center}
	\caption{First and second rows illustrate the reconstruction results from clean measurement data $\y^{0}:=\y$ and noisy measurement data $\y^{\delta}:=\y+\n$ (which contains 10$\%$ Gaussian noise on knee and 5$\%$ Gaussian noise on brain measurement data, respectively). The third raw illustrates the error caused by noisy measurement.}
	\label{f10}
\end{figure}

In this experiment, we test various methods when the measurement data contains larger intensity additional noise. Figure \ref{f10} shows the reconstruction results on the knee and brain dataset using various methods under the calibrated 4x 1-D and 6x 2-D random trajectories with clean and noisy measurement, respectively.
Obviously, for H-Deep-SLR, the reconstructed image has been completely obscured by the noise artifact caused by the noise in measurement. For traditional SPIRiT-POCS algorithm and our H-DEQ-POCS, the noise amplification is significantly lower than H-Deep-SLR. On the knee data, our H-DEQ-POCS outperforms traditional algorithm SPIRiT-POCS, but on brain data, H-DEQ-POCS underperforms SPIRiT-POCS. As for the reasons behind it, we need to further study.

\begin{figure}[thbp]
	\begin{center}
		\subfigure{\includegraphics[width=0.95\textwidth,height=0.48\textwidth]{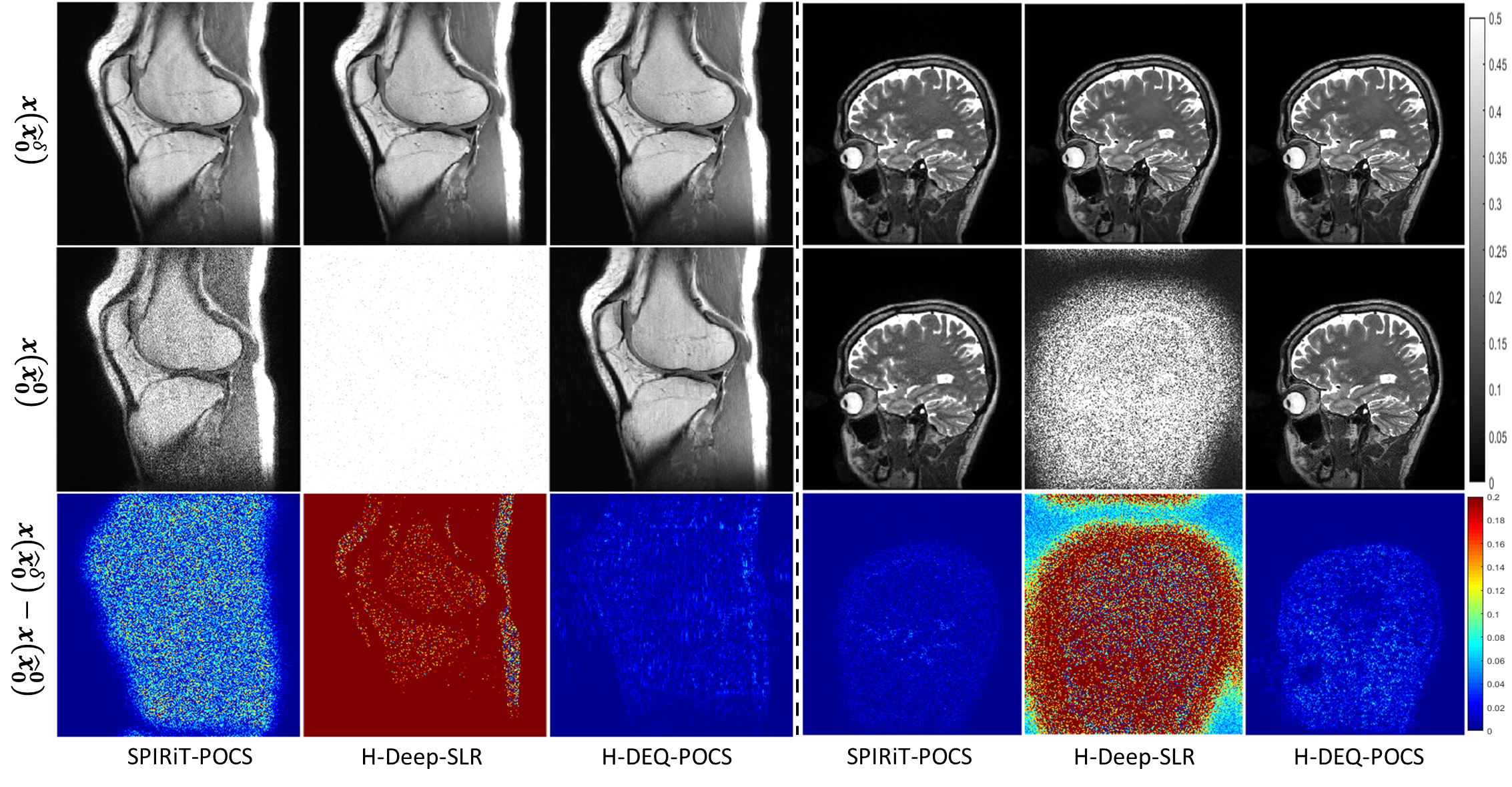}}
	\end{center}
	\caption{First and second rows illustrate the reconstruction results from clean initial input $\widehat{\x}_{0}^{0}:=\widehat{\x}_0$ and noisy initial input $\widehat{\x}_{0}^{\delta}:=\widehat{\x}_0+\n$
		(which contains 200$\%$ Gaussian noise on initial input for knee and 50$\%$ Gaussian noise on initial input for brain data, respectively). The third raw illustrates the error caused by noisy interference on initial input.}
	\label{f11}
\end{figure}

We also test various methods when the initial input contains larger intensity additional noise. Figure \ref{f11} shows the reconstruction results on the knee and brain dataset using various methods under the calibrated 4x 1-D and 6x 2-D random trajectories with clean and noisy initial input, respectively. The reconstructed image by H-Deep-SLR has been also completely obscured by the noise artifact. The influence on the performances caused by noisy initial input is much smaller on
traditional SPIRiT-POCS algorithm and our H-DEQ-POCS. In particular, our H-DEQ-POCS obviously
outperforms traditional algorithm SPIRiT-POCS, which further verifies the robustness of our H-DEQ-POCS.

\subsection{Proof of Theorem \ref{thm:1}}\label{app:2}

	For simplicity, we left out the superscripts here.
	Define $\mathcal{C}=\{\widehat{\x}\in\mathbb{C}^d|\mathcal{M}\widehat{\x}=\y\}$, we have
	\begin{equation*}\begin{aligned}
			&\|\widehat{\x}_{k+1}-\widehat{\x}_{k}\|_F\\
			=&\|\mathcal{P}_{\mathcal{C}}(\Phi_{\phi_{i}}(\widehat{\x}_{k}))-\mathcal{P}_{\mathcal{C}}(\Phi_{\phi_{i}}(\widehat{\x}_{k-1}))\|_F\\
			=&\|(\mathcal{I}-\mathcal{M})( \Phi_{\phi_{i}}(\widehat{\x}_{k})-\Phi_{\phi_{i}}(\widehat{\x}_{k-1}))+\y-\y\|_F\\
			\leq&\|(\mathcal{I}-\mathcal{M})\|\| \Phi_{\phi_{i}}(\widehat{\x}^{\delta}_{k})-\Phi_{\phi_{i}}(\widehat{\x}_{k})\|_F\\
			\leq& L\|\widehat{\x}_{k}-\widehat{\x}_{k-1}\|_F
	\end{aligned}\end{equation*}
	the first equality is due to the definition of $\mathcal{P}_{\mathcal{C}}$ and the last inequality is due to $ \mathcal{I}-\mathcal{M}\preceq\mathcal{I}$ and the $L$-Lipschitz continuity of $\Phi_{\phi_{i}}$. Through recursion on above inequality, we have $$ \|\widehat{\x}_{k+1}-\widehat{\x}_{k}\|_F\leq L^k \|\widehat{\x}_{1}-\widehat{\x}_{0}\|_F.$$
	Summing the above inequality form 0 to $\infty$, we have
	$$ \sum_{k=1}^\infty\|\widehat{\x}_{k+1}-\widehat{\x}_{k}\|_F\leq \frac{1}{1-L} \|\widehat{\x}_{1}-\widehat{\x}_{0}\|_F\leq+\infty$$
	which means that the $\{\widehat{\x}_k\}$ is a Cauchy sequence. Then, it converges to a cluster point globally. The proof is completed.

\subsection{Proof of Theorem \ref{thm:2}}\label{app:3}

	For the fixed point of unrolled POCS algorithm with measurement $\y^m$ and self-consistency $\Phi_{\phi_{KM}}$, i.e., $\mathring{\widehat{\x}}^m=\mathcal{P}_{\{\widehat{\x}|\mathcal{M}\widehat{\x}=\y^m\}}\left(\Phi_{\phi_{KM}}(\mathring{\widehat{\x}}^m)\right)$, $m\in[M]$, we define a sequence of random variables $ \left\{X^m:=\mathring{\widehat{\x}}^m-\widehat{\x}^{m}-\mathbb{E}_{\pi_{\widehat{\x}\times\y}}[\mathring{\widehat{\x}}-\widehat{\x}]\right\}_{m=1}^M$. Since $(\widehat{\x}^m,\y^m)$ is sampled in distribution $\pi_{\widehat{\x}\times\y}$, we have $\mathbb{E}[X^m]=0$ for any $m\in[M]$. Because $\mathring{\widehat{\x}}^m$ is a fixed point, it has to be bounded. Then, there exists a constant $B>0$ such that $\|X^m\|\leq B$. By the Hoeffding's inequality (please see Theorem 7.20 of literature \cite{Foucart2013a} for detail), it holds
	\begin{equation}\label{hoeff1}\begin{aligned}
			&\mathbb{P}\left[\frac{1}{M}\left\|\sum_{m=1}^M\{\mathring{\widehat{\x}}^m-\widehat{\x}^{m}-\mathbb{E}_{\pi_{\widehat{\x}\times\y}}[\mathring{\widehat{\x}}-\widehat{\x}]\}\right\|_F\geq \frac{t_1}{M}\right]\\
			\geq&\mathbb{P}\left[\|\mathbb{E}_{\pi_{\widehat{\x}\times\y}}[\mathring{\widehat{\x}}-\widehat{\x}]\|_F\geq \frac{t_1}{M}+\frac{1}{M}\left\|\sum_{m=1}^M(\mathring{\widehat{\x}}^m-\widehat{\x}^{m})\right\|_F\right]\\
			\geq&\mathbb{P}\left[\|\mathbb{E}_{\pi_{\widehat{\x}\times\y}}[\mathring{\widehat{\x}}-\widehat{\x}]\|_F\geq \frac{t_1}{M}+\frac{1}{M}\sum_{m=1}^M\|\mathring{\widehat{\x}}^m-\widehat{\x}^{m}\|_F\right]\\
			\geq&\mathbb{P}\left[\|\mathbb{E}_{\pi_{\widehat{\x}\times\y}}[\mathring{\widehat{\x}}-\widehat{\x}]\|_F\geq \frac{t_1}{M}+\epsilon\right]\\
	\end{aligned}\end{equation}
	with probability at most $2\exp\left(\frac{-t_1^2}{2MB^2}\right)$, where the first inequality is due to the triangle inequality of Frobenius norm, the second inequality is due to the convexity of Frobenius norm and the last inequality is due to the Assumption \ref{assup:2}. Using the the Hoeffding's inequality again, for the output $\mathring{\widehat{\x}}^n$ of Algorithm \ref{alg:2}, it holds
	\begin{equation}\label{hoeff2}\begin{aligned}
			&\mathbb{P}\left[\|\mathring{\widehat{\x}}^n-\widehat{\x}^{n}-\mathbb{E}_{\pi_{\widehat{\x}\times\y}}[\mathring{\widehat{\x}}-\widehat{\x}]\|_F\geq t_2\right]\\
			\geq&\mathbb{P}\left[\|\mathring{\widehat{\x}}^n-\widehat{\x}^{n}\|_F\geq t_2+\|\mathbb{E}_{\pi_{\widehat{\x}\times\y}}[\mathring{\widehat{\x}}-\widehat{\x}]\|_F\right]\\
	\end{aligned}\end{equation}
	with probability at most $2\exp\left(\frac{-t_2^2}{2B^2}\right)$. Combining inequalities (\ref{hoeff1}) and (\ref{hoeff2}) together, the result is yielded.

\subsection{Proof of Theorem \ref{thm:3}}\label{app:4}

	For simplicity, we left out the superscripts here.
	Let $\widehat{\x}_{k+1}$ denote the unrolled POCS iteration with noise-free measurement, i.e., $\widehat{\x}_{k+1}=\mathcal{P}_{\mathcal{C}}(\Phi_{\phi_{KM}}(\widehat{\x}_k))$ with $\mathcal{C}=\{\widehat{\x}\in\mathbb{C}^d|\mathcal{M}\widehat{\x}=\y\}$, and Let $\widehat{\x}^{\delta}_{k+1}$ denote the unrolled POCS iteration with noisy measurement, i.e.,
	$\widehat{\x}^{\delta}_{k+1}=\mathcal{P}_{\mathcal{C}^\delta}(\Phi_{\phi_{KM}}(\widehat{\x}^{\delta}_{k}))$ with $\mathcal{C}^\delta=\{\widehat{\x}\in\mathbb{C}^d|\mathcal{M}\widehat{\x}=\y^{\delta}\}$. Then, we have
	\begin{equation*}\begin{aligned}
			&\|\widehat{\x}^{\delta}_{k+1}-\widehat{\x}_{k+1}\|_F\\
			=&\|\mathcal{P}_{\mathcal{C}^\delta}(\Phi_{\phi_{KM}}(\widehat{\x}^{\delta}_{k}))-\mathcal{P}_{\mathcal{C}}(\Phi_{\phi_{KM}}(\widehat{\x}_{k}))\|_F\\
			=&\|(\mathcal{I}-\mathcal{M})( \Phi_{\phi_{KM}}(\widehat{\x}^{\delta}_{k})-\Phi_{\phi_{KM}}(\widehat{\x}_{k}))+\y^{\delta}-\y\|_F\\
			\leq&\|(\mathcal{I}-\mathcal{M})\|\| \Phi_{\phi_{KM}}(\widehat{\x}^{\delta}_{k})-\Phi_{\phi_{KM}}(\widehat{\x}_{k})\|_F+\|\y^{\delta}-\y\|_F\\
			\leq& L\|\widehat{\x}^{\delta}_{k}-\widehat{\x}_{k}\|_F + \delta
	\end{aligned}\end{equation*}
	the first equality is due to the definition of $\mathcal{P}_{\mathcal{C}}$ and the last inequality is due to $ \mathcal{I}-\mathcal{M}\preceq\mathcal{I}$ and the $L$-Lipschitz continuity of $\Phi_{\phi_{KM}}$.
	Through recursion on above inequality, we have
	\begin{equation*}
		\|\mathring{\widehat{\x}}^{\delta}-\mathring{\widehat{\x}}\|_F\leq\sum_{k=1}^{\infty} L^{k-1}\delta=\frac{\delta}{1-L}
	\end{equation*}
	where $\mathring{\widehat{\x}}^{\delta}$ and $\mathring{\widehat{\x}}$ denote the fixed points of unrolled POCS iteration with noisy and noise-free measurement, respectively. Combining the result of Theorem \ref{thm:2}, the result is yielded.

\bibliographystyle{unsrtnat}
\bibliography{references}  






\end{document}